\documentclass[10pt,twocolumn,letterpaper]{article}

\usepackage{iccv}
\usepackage{times}
\usepackage{epsfig}
\usepackage{graphicx}
\usepackage{amsmath}
\usepackage{amssymb}

\usepackage{booktabs}
\usepackage{multirow}
\usepackage{pifont}
\usepackage{subfigure}

\usepackage[accsupp]{axessibility} 

\usepackage[pagebackref=true,breaklinks=true,letterpaper=true,colorlinks,bookmarks=false]{hyperref}

\iccvfinalcopy 

\def\iccvPaperID{10723} 

\ificcvfinal\fi

\begin{document}

\newtheorem{definition}{Definition}[section]
\newtheorem{claim}{Claim}[section]
\newtheorem{proposition}{Proposition}
\title{ModelGiF: Gradient Fields for Model Functional Distance}

\author{Jie Song, Zhengqi Xu, Sai Wu, Gang Chen, and Mingli Song\thanks{Corresponding author}\\
Zhejiang Univerisy\\
{\tt\small \{sjie,xuzhengqi,wusai,cg,brooksong\}@zju.edu.cn}}

\maketitle
\ificcvfinal\thispagestyle{empty}\fi

\begin{abstract}
   The last decade has witnessed the success of deep learning and the surge of publicly released trained models, which necessitates the quantification of the model functional distance for various purposes. However, quantifying the model functional distance is always challenging due to the opacity in inner workings and the heterogeneity in architectures or tasks.
   Inspired by the concept of ``field'' in physics, in this work we introduce \textbf{Model Gradient Field}~(abbr. ModelGiF) to extract homogeneous representations from the heterogeneous pre-trained models. Our main assumption underlying ModelGiF is that each pre-trained deep model uniquely determines a ModelGiF over the input space. The distance between models can thus be measured by the similarity between their ModelGiFs. We validate the effectiveness of the proposed ModelGiF with a suite of testbeds, including task relatedness estimation, intellectual property protection, and model unlearning verification. Experimental results demonstrate the versatility of the proposed ModelGiF on these tasks, with significantly superiority performance to state-of-the-art competitors. Codes are available at \url{https://github.com/zju-vipa/modelgif}.
\end{abstract}

\section{Introduction}
\label{sec:intro}
The last decade has witnessed the great progress of deep learning in various fields, and a plethora of deep neural networks are developed and released publicly, with either their architectures and trained parameters~(\eg, Tensorflow Hub\footnote{\url{https://www.tensorflow.org/hub}}, Pytorch Hub\footnote{\url{https://pytorch.org/hub/}}) for research, or the prediction API~(\eg, BigML, Amazon Machine Learning) as ML-as-a-Service~(MLaaS) for commercial purposes. These off-the-shelf pre-trained models become extremely important resources for not only practitioners to solve their own problems, but also researchers to explore and exploit the huge potential underlying these pre-trained models.

\begin{figure}[t]
  \centering
    \includegraphics[width=1.0\linewidth]{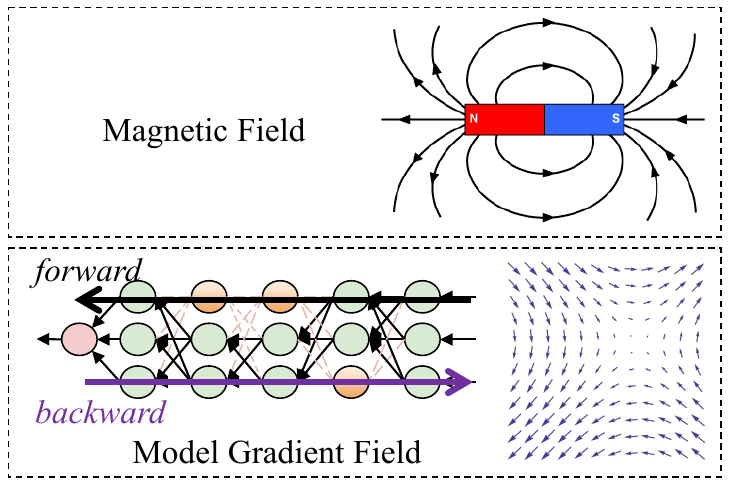}
  \caption{An illustrative diagram of the \textit{magnetic field} and the proposed \textit{model gradient field}~(ModelGiF) defined on the input space.}
  \label{fig:field}
  \vspace{-1.5em}
\end{figure}

With the surge of pre-trained models released, quantifying the model relationship  emerges as an important question. The large-scale open-sourced deep models, heterogeneous in architectures and tasks and trained isolatedly or dependently, are related to each other in various manners. For example, a student model trained by knowledge distillation~\cite{hinton2015distilling} should behave more similarly with the teacher than an independently trained identical model. Likewise, a fine-tuned model should be more closely related to its pre-trained model than the one trained from scratch. More generally, models trained in isolation on heterogeneous tasks should inherit the intrinsic task relatedness~\cite{zamir2018taskonomy} as task-specific features are extracted by these models. Broadly speaking, there exists a \textit{model metric space} where models with similar functional behaviors are clustered together and dissimilar ones are far apart. The model functional distance between models, if left unresolved, leaves existing model repositories still simple unstructured collections of many isolated open-sourced models, hindering the exploitation of their great value as a whole.

Despite the ever-increasing number of publicly available pre-trained models, the study on the model functional distance, \ie, structure of the model metric space, lags far behind. This phenomenon can be largely attributed to the great challenge of computing the functional distance between any deep models, where the barriers are three-folds: (1)~\textit{Heterogeneity}: deep models usually differ significantly in architectures. Even with the same architecture, models trained on different datasets or tasks can also behave quite differently; (2) \textit{Opacity}: the opaque inner workings of deep models render computing the model similarity extremely difficult; (3) \textit{Efficiency}: as the cost of computing pairwise distance grows quadratically with the number of models, the computation of the functional distance should be efficient.

A few prior works have been devoted to computing model distance, in either weight space~\cite{achille2019task2vec,jia2021proof} or representation space~\cite{li2021modeldiff,dwivedi2019representation}. For example, Task2Vec~\cite{achille2019task2vec} computes task or model representations based on estimates of the Fisher information matrix associated with the probe network parameters, which provides a fixed-dimensional embedding of the model agnostic to the task variations. However, Task2Vec assumes all probe models share the same architecture, which is highly restrictive in practice. ModelDiff~\cite{li2021modeldiff} and representation similarity analysis~(RSA)~\cite{dwivedi2019representation}, on the other hand, adopt the similarity of representations on a small set of reference inputs as the model representations, and the model distance is thus calculated resorting to these model representations, which is shown effective for model reuse detection and transferability estimation. However, these methods only capture the point-wise behavior of the models at a small number of chosen reference points, which is limited in representation capacity and fails at harder tasks such as detecting model extraction~\cite{li2021modeldiff}. Recently, DEPARA~\cite{song2019deep,song2020depara} and ZEST~\cite{jia2021zest} are proposed as pilot studies by applying explanation methods for computing model distance. However, they are validated on disjoint downstream tasks, and the performance on comprehensive tasks remains unclear. Moreover, these methods also use a small number of chosen reference points to extract the model representations, which makes them suffer from low representation capacity. 

In this work, inspired by the concept of ``field'' in physics, we propose \textbf{Model Gradient Field}~(as shown in Figure~\ref{fig:field}), abbreviated as ModelGiF, as the proxy to extract homogeneous representations from pre-trained heterogeneous models to derive their functionality distance. Specially, the proposed ModelGiF is defined on the input space, \ie, every point in the ModelGiF denotes the gradient vector of the model output \textit{w.r.t.} the input on the same point. The main assumption underlying ModelGiF is that each pre-trained deep model uniquely determines a ModelGiF over the input space. The functional distance between any two models can thus be measured by the similarity between their ModelGiFs. Unlike prior methods where the point-wise features are adopted for representing the model, ModelGiFs represents these models by their gradient on the whole input space, which makes it more capable of differentiating highly related models~(\ie, model extraction detection). Moreover, we provide theoretical insights into the proposed ModelGiFs for model functional distance, and make extensive discussions on different implementation details. We validate the effectiveness of the proposed ModelGiF with a suite of testbeds, including transferability estimation, intellectual property protection, and model unlearning verification. Experimental results demonstrate the versatility of the proposed ModelGiF on these tasks, with significantly superiority to state-of-the-art~(SOTA) methods.

To sum up, we make the following contributions: (1) we propose the concept of ``model gradient field'', a novel method for quantifying the functionality similarity between pre-trained deep models; (2) we provide theoretical insights into the proposed ModelGiFs for model functional distance, and make extensive discussions on different implementation details; (3) extensive experiments demonstrate the effectiveness and the superiority of ModelGiF on various tasks, including transferability estimation, intellectual property protection, and model unlearning verification.

\section{Related Work}
This section briefly reviews some related topics to model functional distance, including task relatedness estimation, intellectual property protection, and model unlearning.

\vspace{0.5em}
\noindent \textbf{Task Relatedness Estimation.}
Recent studies~\cite{zamir2018taskonomy,song2019deep,song2020depara,bao2019information,yosinski2014transferable,azizpour2015factors,ben2003exploiting, kifer2004detecting} reveals the existence of task relatedness or task structure among visual tasks. It is the concept underlying transfer learning and provides a principled way  to seamlessly reuse supervision among related tasks or solve many tasks in one system~\cite{zamir2018taskonomy}. Existing works to obtain task relatedness can be roughly divided into empirical and analytical approaches. Taskonomy~\cite{zamir2018taskonomy} is the most representative work of empirical methods. It proposes a fully computational approach to obtain task relatedness by exhaustively computing the actual transfer learning performance, which is thus usually taken as the ground truth for evaluating other estimators. Despite the indisputable results, the computational cost of empirical methods is extremely high. On the other hand, analytical methods, \textit{e.g.,}, DEPARA~\cite{song2020depara,song2019deep}, Task2Vec~\cite{achille2019task2vec} and RSA~\cite{dwivedi2019representation}, try to estimate the task relatedness without conducting the actual transfer learning. Analytical methods generally significantly reduce the computation overhead, but can be highly restrictive in model architectures or estimation performance. The proposed ModelGiF relaxes these restrictions and achieves task relatedness more consistent to taskonomy than existing approaches.

\vspace{0.5em}
\noindent \textbf{Intellectual Property Protection.}
Since training deep models usually consumes expensive data collection and large computation resources, the trained models constitute valuable~\textit{intellectual property~(IP)} that should be protected in cases where reusing is not allowed without authorization. However, model IP can be infringed 
in a variety of forms, such as fine-tuning~\cite{adi2018turning,uchida2017embedding}, pruning~\cite{liu2018fine,molchanov2019importance}, transfer learning~\cite{sharif2014cnn, weiss2016survey,ying2018transfer} and model extraction~\cite{orekondy2019knockoff,tramer2016stealing, juuti2019prada,jagielski2020high}, which poses great challenge to IP protection. Existing IP protection approaches can be roughly categorized into~\textit{watermarking}~\cite{jia2021entangled,uchida2017embedding,fan2019rethinking,zhang2018protecting,adi2018turning} and~\textit{fingerprinting}~\cite{lukas2019deep,cao2021ipguard,li2021modeldiff,guan2022you}. Watermarking methods
usually leverage weight regularization~\cite{uchida2017embedding,fan2019rethinking} to put secret watermark in the model parameters or train models on a triggered set to leave backdoor~\cite{adi2018turning,zhang2018protecting} in them. While being able to provide exact ownership verification, these techniques are invasive, \ie, they need to tamper with the training process, which may affect the model utility or introduce new security risks into the model. Fingerprinting, on the contrary, extracts
a unique identifier, \ie, fingerprint, from the owner model to differentiate it from other models. Latest fingerprinting (\eg, ModelDiff~\cite{li2021modeldiff}, IPGuard~\cite{cao2021ipguard}, SAC~\cite{guan2022you}) on deep learning models, though being non-invasive, also fall short when facing the diverse and ever-growing attack scenarios~\cite{chen2022copy}. In this work, we propose ModelGiF as the homogeneous representations for heterogeneous pre-trained models, which can be naturally adopted as the fingerprint for IP protection and yields superior IP protection performance to existing approaches.

\vspace{0.5em}
\noindent\textbf{Model Unlearning Verification.}
Model unlearning aims to remove the effects of data points from the trained model, which has been attracting increasing attentions in recent years due to legislation and privacy ethics. Cao~\textit{et al.}~\cite{cao2015towards} are dedicated to making learning systems forget and present an unlearning approach capable of forgetting specific data. Graves \textit{et al.}~\cite{graves2021amnesiac} propose an effective method to eliminate the influence of personal data from the model while maintaining the validity. Bourtoule \textit{et al.}~\cite{bourtoule2021machine} particularly propose a model unlearning method to decrease the time it takes to retrain exactly. However, how to identify whether the impact of data has been eliminated is an essential but rarely studied problem. Recently ZEST~\cite{jia2021zest} verifies the unlearning of data by comparing the Local Interpretable Model-Agnostic Explanations (LIME)~\cite{ribeiro2016should}. In this work, we adopt the proposed ModelGiF for unlearning verifacation, which yields competitive performance in our experiments.

\section{Methodology}
\subsection{Problem Setup} 
\label{sec:problem-setup}
Assume there is a model repository consisting of $N$ pre-trained models $\mathcal{M}=\left \{{M_{1}, M_{2}, ..., M_{N}}\right \}$. With mild assumptions, these models are defined on the same input space $\mathbb{R}^{D}$ ($D=WHC$, where $W$, $H$ and $C$ denotes the width, height and channel of the input space\footnote{Without losing generality, we use vector instead of tensor for notation simplicity.}), yet trained with data sampled from different data distributions $\mathcal{P}=\left\{p_1, p_2, ..., p_N\right\}$. Note that here we made no assumptions on the model architectures and the tasks, which means that these models can be different in architectures~(\eg, ResNet~\cite{he2016deep} and VGG~\cite{simonyan2014very}), and tasks~(\eg, visual classification~\cite{krizhevsky2017imagenet}, detection~\cite{ren2015faster} or segmentation~\cite{chen2017rethinking}). The goal in this work is to construct a model metric space where models with similar functional
behaviors are clustered together and dissimilar ones are far apart, where the vital step is quantifying the functional distance between these models. 

\begin{figure*}[htbp]
  \centering
   \includegraphics[scale=0.62]{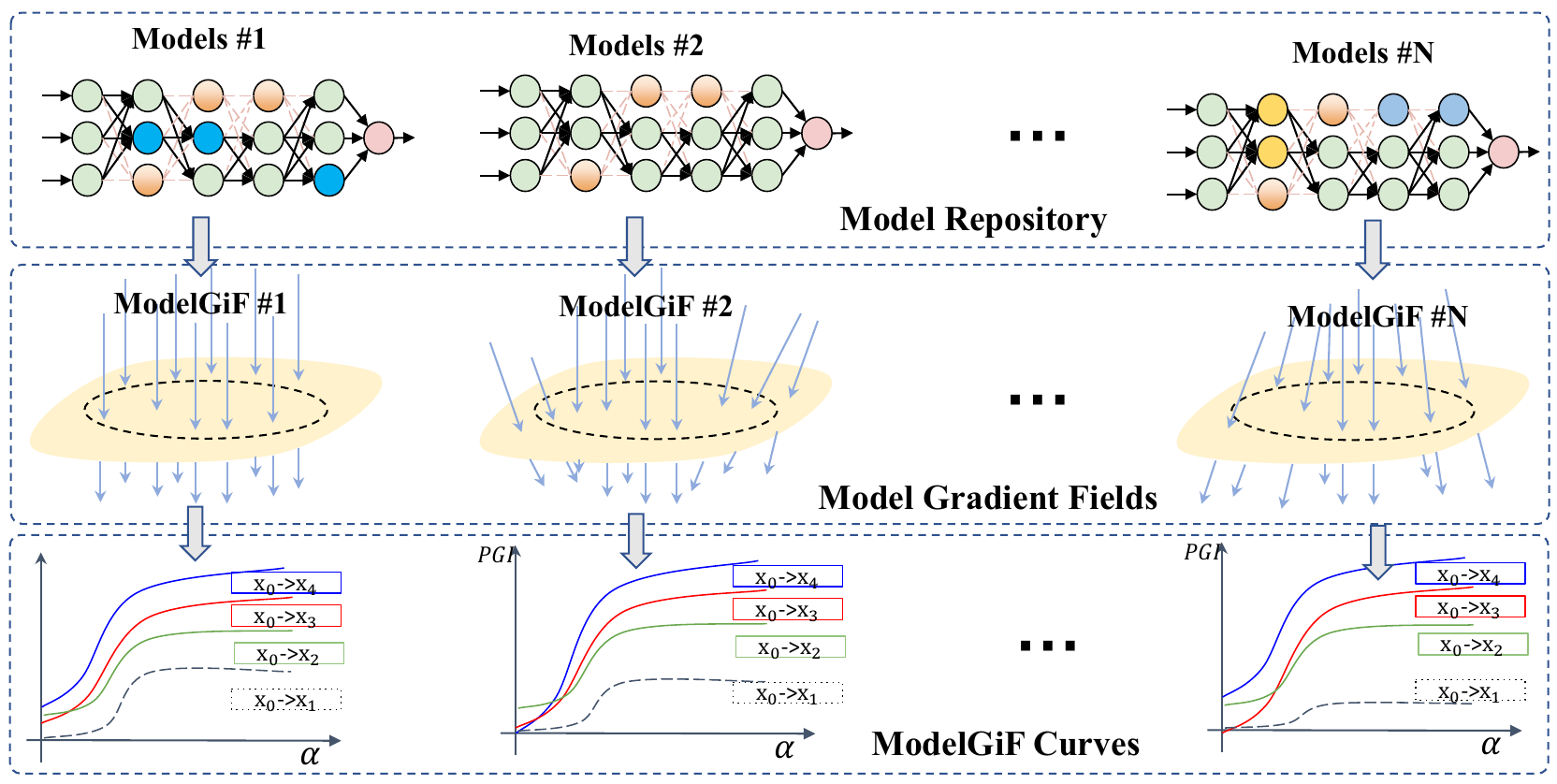}
   \caption{An illustrative diagram of the overall pipeline of obtaining ModelGiF curves. The more similar the ModelGiFs, the more similar the corresponding trained models. Note that ModelGiF Curves are in high-dimensional space in reality. }
   \label{fig:overview}
\end{figure*}

\subsection{The Proposed ModelGiF}
\label{sec:gif}
As aforementioned, quantifying the functional distance between pre-trained models is challenging due to the heterogeneity, opacity and efficiency issues.
Our main idea is extracting homogeneous descriptors from these heterogeneous models to make them comparable. To this end, we proposed the concept ModelGiF, as shown in Figure~\ref{fig:field}, to derive the comparable identifier of the pre-trained models for quantifying model distance.

ModelGiF is inspired by the concept ``field'' in physics where a field is a mapping from space to some physical quantity, represented by a scalar, vector, or tensor, that has a value for each point in the space~\cite{mcmullin2002origins}. For example, the magnetic field describes the magnetic influence on moving electric charges, electric currents, and magnetic materials~\cite{Feynman1965TheFL}. Likewise, we define ModelGiF as a mapping from the input space to the gradient space such that each point in the input space is assigned to a gradient vector. The formal definition is provided as follows.
\begin{definition}[Model Gradient Field]
Let $M$ be a deep model trained on the data which is sampled according to some distribution $p$ from the data space $\mathcal{X}$, outputting a scalar\footnote{For vector or tensor predictions, we simply take their $l_2$ norm as the scalar output.} prediction in the label space $\mathcal{Y}$. A point $\mathbf{x}$ can be described as $\mathbf{x}=(x_1, x_2, ..., x_D)$. We define the model gradient field of $M$ as the gradient of every possible input point in $\mathcal{X}$ w.r.t. the model output:
\begin{equation}
    \textsc{ModelGiF}(M)\triangleq\nabla_{\mathbf{x}}M,
\end{equation}
\end{definition}
which is a mapping from the input space $\mathbb{R}^D$ to the gradient space $\mathbb{G}^D$, $\textsc{ModelGiF}:\mathbb{R}^D\rightarrow \mathbb{G}^D$. Note that the ModelGiF is defined on the whole input space~(which is shared by all the models) rather than some discretely points sampled from the data distribution $p$ specific to the model $M$. As the architecture of $M$ and the tasks~(\ie, the label space $\mathcal{Y}$ ) can be heterogeneous, the ModelGiF of a model can be seen as a projection of the model to the common input space.

\subsection{Field Curves for ModelGiF}
With the proposed ModelGiF, the model functional distance can be quantified by the ModelGiF similarity. However, as ModelGiF is defined on the whole input space that is usually extremely huge, how to extract its representation or signature becomes the vital step for quantifying model distance. In physics, a field is usually depicted by the \textit{field curves}~(as shown in Figure~\ref{fig:field}), \ie, the integral curves for the field, and can be constructed by starting at a point and tracing a curve through space that follows the direction of the vector field, by making the field curve tangent to the field vector at each point~\cite{theisel1995vector}. Inspired by this, we propose \textit{ModelGiF Curves} as the descriptors of ModelGiF to measure the model functional distance.
\begin{definition}[Field Curves]
\label{def:fcurve}
For a field F: $\mathbb{R}^{D}\rightarrow\mathbb{G}^D$, a curve $\mathbf{x}(t)=\big(x_1(t),x_2(t),...,x_D(t)\big)$ is called a field curve of the field F if the following condition is satisfied: For all points $P\in \mathbf{x}(t)$, the tangent vector of the curve in the point $P$ has the same direction as the vector $F(P)$:
\begin{equation}
\label{eq:field-curve}
    \frac{dx_1(t)}{dt}=F\big(\mathbf{x}(t)\big)_1, \frac{dx_2(t)}{dt}=F\big(\mathbf{x}(t)\big)_2, ...
\end{equation}
\end{definition}

The field curve can be described as the solution of the system of differential equations in Eqn.~\ref{eq:field-curve}, and it plays an important role for both analysis and visualization of vector fields~\cite{theisel1995vector}. Unfortunately, as the ModelGiF is usually complicated for a pre-trained deep model, there is no closed solution. The field curves are in general not describable as parameterized curves, which hinders the comparisons between different ModelGiFs. To resolve this issue, we introduce the definition of ModelGiF curves as follows.
\begin{definition}[ModelGiF Curves]
\label{def:mcurve}
For a ModelGiF F: $\mathbb{R}^{D}\rightarrow\mathbb{G}^D$, a curve $\mathbf{g}(t)=\big(g_1\left(t\right), g_2\left(t\right), ..., g_D\left(t\right)\big)$ is called the ModelGiF curve of F if the following condition is satisfied: 
\begin{equation}
    \frac{dg_1(t)}{dt}=F(\mathbf{x}(t))_1, \frac{dg_2(t)}{dt}=F(\mathbf{x}(t))_2, ...
\end{equation}
\end{definition}
Note that different from the definition of field curves, the ModelGiF curves are defined on the gradient space $\mathbb{G}^D$ instead of the input space $\mathbb{R}^D$. 
\begin{proposition}
\label{pro:modelgif-curve}
For a ModelGiF F: $\mathbb{R}^{D}\rightarrow\mathbb{G}^D$, and two points $\mathbf{x}_0$, $\mathbf{x}_1$ in $\mathbb{R}^D$, the gradient integral along the straight line from $\mathbf{x}_0$ to $\mathbf{x}_1$ is a ModelGiF curve of F: 
\begin{equation}
\label{eq:prop-gif-curve}
    \mathbf{g}(t) = \int_{\alpha=0}^{t}F\big(\mathbf{x}_0+\alpha(\mathbf{x}_1-\mathbf{x}_0)\big)d\alpha, \ \ \ t\in[0, 1]
\end{equation}
\end{proposition}
With the proposition, we can make comparisons between ModelGiFs by simply comparing their ModelGiF curves between some predefined pairs of points. Now we provide the detailed pipeline of quantifying the Model functional distance with the proposed ModelGiF. 

\subsection{ModelGiF for Model Distance}
Provided with the trained deep models as described in Section~\ref{sec:problem-setup}, we compute the model distance with the proposed ModelGiF in two steps~(in Figure~\ref{fig:overview}): \textit{1) Sampling reference points}, and \textit{2) Computing the model distance}.  

\vspace{0.5em}
\noindent \textbf{Sampling Reference Points.}
As described in Proposition~\ref{pro:modelgif-curve}, the first step before obtaining ModelGiF curves is determining the reference points~(\ie., $\mathbf{x}_1$\footnote{Here we simply set $\mathbf{x}_0$ to the zero vector. Other settings do not yield significantly superior performance in our experiments.} in Eqn.~\ref{eq:prop-gif-curve}.), which determines the domain of these curves. Let $K$ be the number of reference points to be sampled. Intuitively, these reference points should \textit{representative}. However, as the models in $\mathcal{M}$ can be heterogeneous in architectures and tasks and trained on data sampled from different distributions, it is hard to determine a common set of reference points which are representative for all these models. In this work, we investigate three types of reference points as follows. 

1) \textit{Random samples} drawn from $\mathcal{P}$. Each data point is sampled in two stages: first randomly sampling the distribution $p$ from $\mathcal{P}$ and then randomly sampling the reference point from $p$.

2) \textit{Augmented samples} using CutMix~\cite{yun2019cutmix}. Let $\mathbf{x}$ and $\mathbf{x}^{*}$ be two randomly sampled points in $\mathcal{P}$.  CutMix generates a new sample $\Tilde{\mathbf{x}}$ using the two samples by cutting off and pasting patches among them: $\Tilde{\mathbf{x}}=\mathbf{m}\odot\mathbf{x}+(1-\mathbf{m})\odot\mathbf{x}^*$, where $\mathbf{m}$ represents the binary mask to combine images with different parts, and $\odot$ the element-wise multiplication.

3) \textit{Adversarial samples} generated with PGD~\cite{madry2017towards}. PGD attack is a multi-step variant of  Fast Gradient Sign Method~(FGSM)~\cite{goodfellow2014explaining}: $\mathbf{x}_{t+1} = \prod\big(\mathbf{x}_t + \alpha \textsc{sgn}(\nabla \mathcal{J}(\mathbf{x}_t))\big)$, where $\prod$ denotes the projection to the allowed space. 

The performance of these types of reference points are discussed in Section~\ref{sec:exp-task-relatedness}. We also provide the sensitivity analysis of the number of reference points $K$, which demonstrates that the performance of ModelGiF quickly becomes superior to existing methods as $K$ increases.

\vspace{0.5em}
\noindent \textbf{Computing the model distance.}
Let $\mathcal{S}=\{\mathbf{x}_1, \mathbf{x}_2, ,\mathbf{x}_K\}$ be the reference points obtained from the last step. For the $i$-th model $M_i$, we can get $K$ ModelGiF curves by substituting each point in $\mathcal{S}$ for the $\mathbf{x}_1$ in Eqn.~\ref{eq:prop-gif-curve}. These curves are used as the identifier of the ModelGiF, and thus serve as the representations of the trained deep model. There are several distance metrics to measure the similarity between curves, \eg, Hausdorff distance and Fréchet distance. As the method for curve distance is not our contribution in this work, we simply adopt the integrated point-wise cosine distance to validate the performance of the proposed method:
\begin{equation}
\label{eq:curve-distance}
    d(M_i, M_j) = \sum_{k=1}^{K}\int_{t=0}^{1}\big(1-\textsc{cos}(\mathbf{g}^{i,k}(t),  \mathbf{g}^{j,k}(t))\big)dt,
\end{equation}
where $\mathbf{g}^{i,k}$ denotes the $k$-th curve of the $i$-th model in $\mathcal{M}$, and $\cdot$ denotes the inner product between $\mathbf{g}^{i,k}(t)$ and $\mathbf{g}^{i,k}(t)$. In practice, we approximate the integral with summation by sampling with some fixed intervals.

\subsection{Theoretical Analysis of ModelGiF}
From the definition of model distance in Eqn.~\ref{eq:curve-distance}, we can see that the defined distance can be seen as the summation of point-wise similarities over all the curves. Here we provide some theoretical insights into ModelGiF by  dissecting the model-level distance  into point-level distance.  

\begin{proposition}
\label{pro:theory}
For the point $\mathbf{g}(t_1)$ along the ModelGiF curves $\mathbf{g}(t)$ defined from $\mathbf{x}^*$ to $\mathbf{x}$, then
\begin{equation}
\label{eq:ig}
 \sum\nolimits_i t_1(x_{i}-x^*_i){g_i(t_1)} = M(\mathbf{x}) - M(\mathbf{x}^*),
\end{equation}
\end{proposition}
where the left side is actually the summation over Integrated Gradients~\cite{sundararajan2017axiomatic}. Eqn.~\ref{eq:ig} tells us that the every point in proposed ModelGiF curve is strongly related to the model prediction differences from this point and the baseline point, thus the proposed model distance is a powerful distance metric to quantifying the model function distance.

\begin{figure*}[htbp]
  \centering
  \subfigure[Affinity Matrix from Taskonomy]{
    \begin{minipage}{0.32\linewidth}
    \centering   
    \includegraphics[scale=0.38]{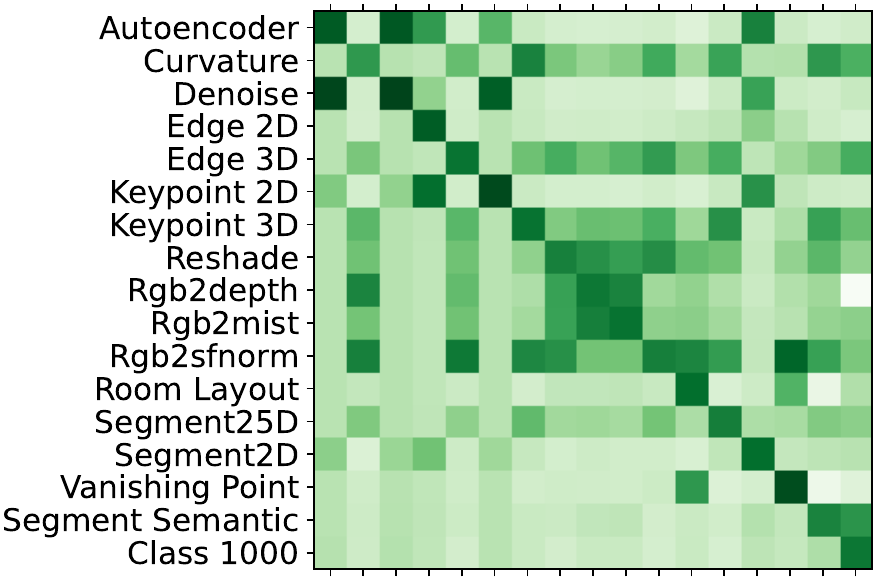}  
    \end{minipage}}
    \subfigure[Affinity Matrix from RSA]{
    \begin{minipage}{0.32\linewidth}
    \centering   
    \includegraphics[scale=0.38]{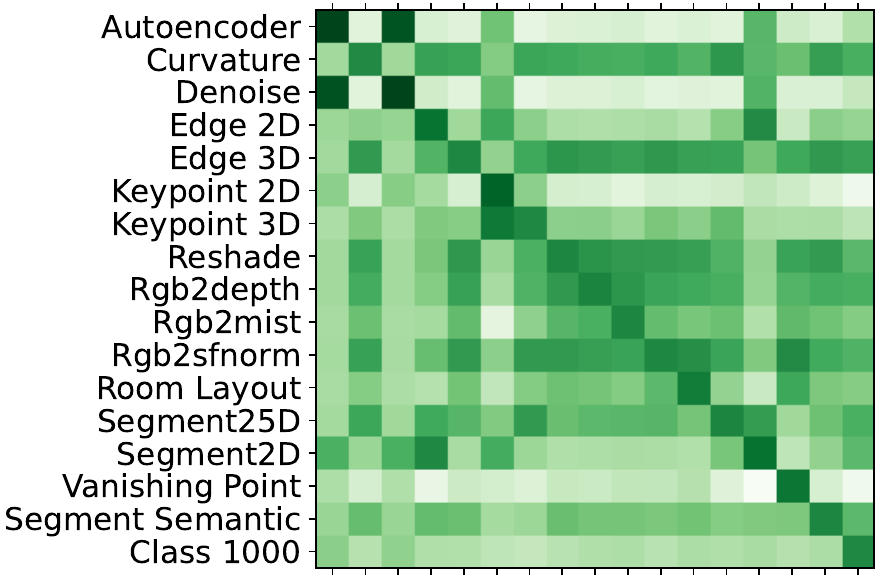}  
    \end{minipage}}
    \subfigure[Affinity Matrix from ModelGiF]{
    \begin{minipage}{0.32\linewidth}
    \centering   
    \includegraphics[scale=0.38]{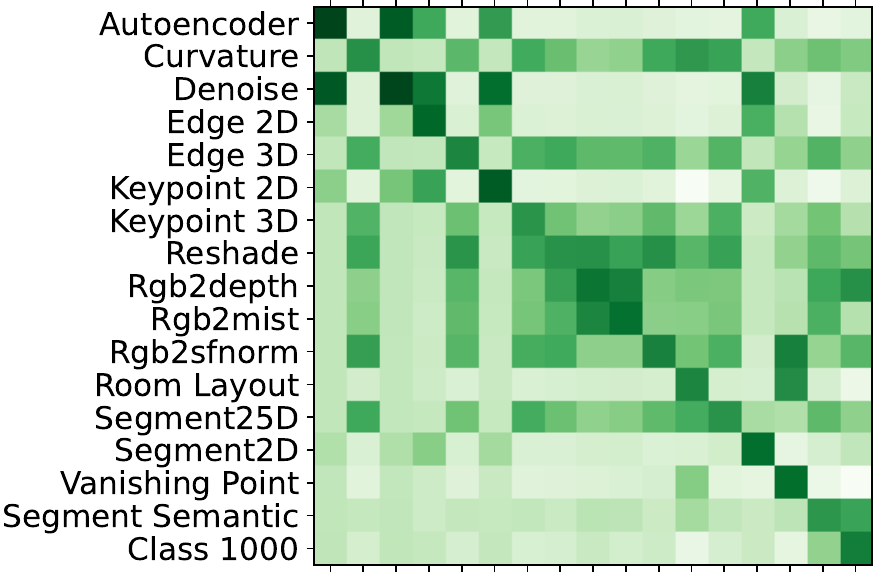}  
    \end{minipage}}\\
    \subfigure[Similarity Tree from ModelGiF]{
    \begin{minipage}{0.32\linewidth}
    \centering   
    \includegraphics[height=3.5cm]{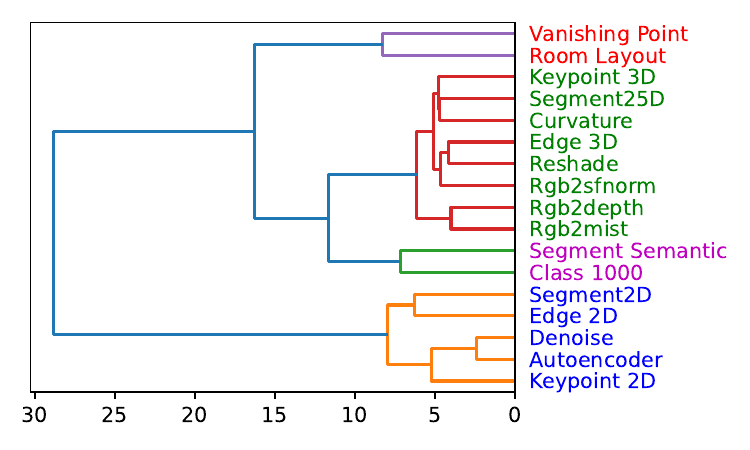}  
    \end{minipage}}
    \subfigure[Different reference points]{
    \begin{minipage}{0.32\linewidth}
    \centering   
    \includegraphics[height=3.5cm]{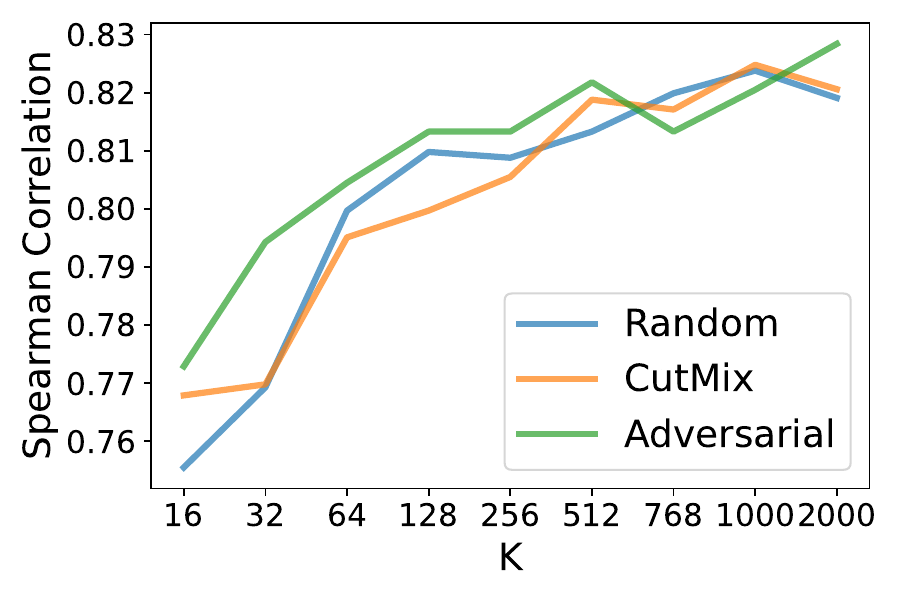}  
    \end{minipage}}
    \subfigure[Different implementations]{
    \begin{minipage}{0.32\linewidth}
    \centering   
    \includegraphics[height=3.5cm]{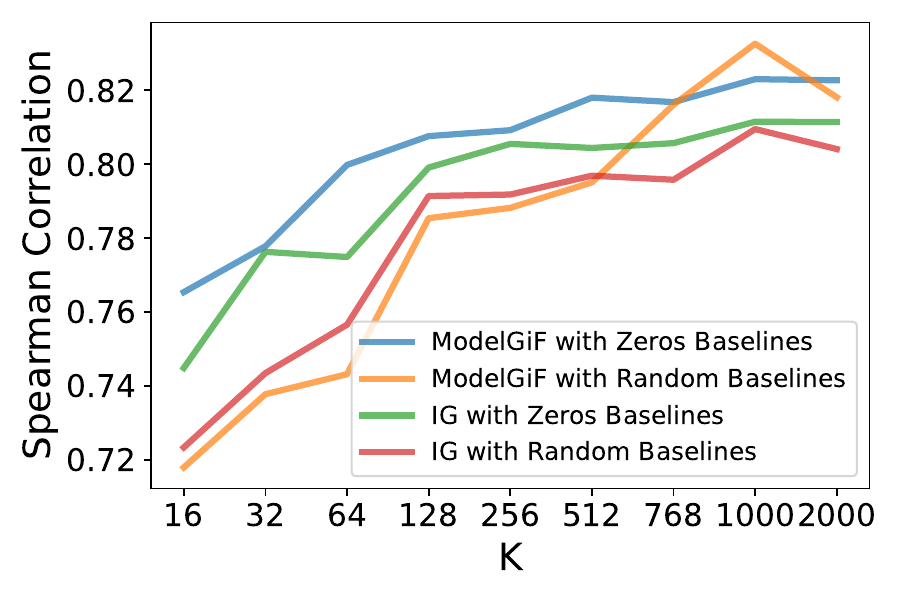}  
    \end{minipage}}
  \caption{ Results of task related estimation:
  (a) affinity matrix from Taskonomy; (b) affinity matrix from RSA; (c) affinity matrix from ModelGiF; (d) the similarity tree derived from ModelGiF; (e) performance with varying $K$ of different reference points; (e) performance with varying $K$ of different implementations.}
  \label{fig:taskonomy}
\end{figure*}

\section{Experiments}
In this section, we verify the effectiveness of ModelGiF with a suite of testbeds, including task relatedness estimation,  intellectual property protection and model unlearning verification. Details are provided as follows.

\subsection{Application: Task Relatedness Estimation}
\label{sec:exp-task-relatedness}
\vspace{0.5em}
\noindent \textbf{Experimental Setup.}
We adopt 17 pre-trained heterogeneous models from Taskonomy~\cite{zamir2018taskonomy} to compare their functionality similarity. These models are trained on various tasks~(including Autoencoder, Curvature, Denoise, Edge 2D, Edge 3D, Keypoint 2D, Keypoint 3D, Reshade, Rgb2depth, Rgb2mist, Rgb2sfnorm, Room Layout, Segment25D, Segment2D, Vanishing Point, Segmentation, and Classification) and are different in architectures. Generally speaking, the architectures of these models follows an encoder-decoder scheme, in which the encoder is implemented by fully convolutional layers and the decoder varies according to the tasks. Please refer to~\cite{zamir2018taskonomy} for more detailed information. In our experiments, only the encoders of these models are adopted for generating their ModelGiFs. 

\vspace{0.5em}
\noindent \textbf{Competitors and Evaluation Metric.} The proposed ModelGiF is compared with several prior approaches to task relatedness estimation, including RSA~\cite{dwivedi2019representation},  AttributionMaps~\cite{song2019deep}, and ZEST~\cite{jia2021zest}. Note that AttributionMaps adopts saliency~\cite{simonyan2013deep}, DeepLIFT~\cite{shrikumar2016not} and $\epsilon$-LRP~\cite{bach2015pixel} to generate attributions for task relatedness estimation. We compare ModelGiF with all these variants  to demonstrate its superiority in task relatedness prediction.
As the affinity matrix  from Taskonomy is obtained based on the actual transfer learning performance, it is used as the ground truth of the affinity among these tasks.
In this experiment, we will verify that the functionality similarity obtained by ModelGiF positively correlates with that from taskonomy. In order to quantify the similarity between the affinity matrix representing functionality similarity obtained by ModelGiF and the affinity matrix obtained by Taskonomy, we use Spearman's correlation as the metric for evaluation. 

\begin{table}[t]
  \centering
\begin{tabular}{lc}
\toprule
\textbf{Method} & \textbf{Spearman's Correlation} \\
\midrule
Zest~\cite{jia2021zest} & 0.359 \\
AttributionMaps$_{\textsc{Saliency}}$~\cite{song2019deep} & 0.619 \\
AttributionMaps$_{\textsc{DeepLIFT}}$~\cite{song2019deep} & 0.685 \\
AttributionMaps$_{\epsilon\textsc{-LRP}}$~\cite{song2019deep} & 0.682 \\
RSA~\cite{dwivedi2019representation} & 0.777 \\
\hline
ModelGiF$_{\textsc{Random}}$ & {0.834}\\
ModelGiF$_{\textsc{Augment}}$ & {0.835}\\
ModelGiF$_{\textsc{Adversarial}}$ & {0.830}\\
\bottomrule
\end{tabular}
  \caption{Comparison between the proposed ModelGiF and existing methods. Here $1,000$ reference points are sampled in ModelGiF.}
  \label{tab:att}
\end{table}

\vspace{0.5em}
\noindent \textbf{Results.} We first test the proposed ModelGiF with by randomly sampling $1,000$ reference points. The visualization of the affinity matrix (as well as that from Taskonomy~\cite{zamir2018taskonomy} and RSA~\cite{dwivedi2019representation}) and the similarity tree are provided in Figure~\ref{fig:taskonomy}. This similarity tree is constructed by agglomerative hierarchical clustering based on the affinity matrix. It can be seen that the affinity matrix from the proposed ModelGiF is visually highly similar to that from Taskonomy in most regions. The similarity tree derived from ModelGiF perfectly matches the results from taskonomy where 3D~(in green font) tasks, 2D~(in blue font), geometric~(in red font) tasks, and semantic~(in purple font) tasks cluster into corresponding groups as expected.

\begin{table*}[t]
  \centering
\begin{tabular}{l|ccccc|ccccc}
\toprule
&\multicolumn{5}{c|}{\textbf{CIFAR}} &\multicolumn{5}{c}{\textbf{tiny-ImageNet}}\\
\midrule
\multirow{2}{*}{\textbf{Attack}} & IPGuard & CAE & EWE & SAC & ModelGiF & IPGuard & CAE & EWE & SAC & ModelGiF\\
& \cite{cao2021ipguard} & \cite{lukas2019deep} & \cite{jia2021entangled}& \cite{guan2022you} & (Ours)& \cite{cao2021ipguard} & \cite{lukas2019deep} & \cite{jia2021entangled}& \cite{guan2022you} & (Ours)\\
\midrule
Finetune-A & 1.00 & 1.00 & 1.00 & 1.00& 1.00& 1.00 &1.00 &0.48 &1.00 & 1.00\\
Finetune-L & 1.00 & 1.00 & 1.00 & 1.00& 1.00& 1.00 &1.00 &1.00 &1.00 & 1.00\\
Pruning & 1.00 & 0.95 & 0.87 & 1.00 & 1.00& 1.00 &1.00 &0.58 & 1.00 & 1.00\\
Extract-L & 0.81 & 0.83 & 0.97 & 1.00& 1.00& 0.97&1.00 &1.00 & 1.00 & 1.00\\
Extract-P & 0.81 & 0.90 & 0.97 & 1.00& 1.00& 0.97&1.00 & 1.00 & 1.00 & 1.00\\
Extract-Adv & 0.54 & 0.52 & 0.91 &0.92 &1.00 & 0.65&0.78 & 1.00 &0.91 & 1.00\\
Transfer-10C & 1.00 & 1.00 & 1.00 & 1.00 & 1.00 & N/A  & N/A & N/A & N/A & N/A\\
Transfer-A & -- & -- & -- &1.00 &1.00 &-- &-- & --& 1.00 &1.00 \\
Transfer-L & -- & -- & --&1.00 &1.00 &-- & --& --& 1.00 &1.00 \\
\bottomrule
\end{tabular}
  \caption{Comparison between the proposed ModelGiF and existing methods for IP proctection. The performance is measured in AUC value under the AUC-ROC curve. ``--'' represent the IP protection method can not detect this kind of attack. ``N/A'' denotes not applicable.}
  \label{tab:ip-protection}
\end{table*}

Table~\ref{tab:att} provide a quantitative comparison between the proposed ModelGiF with existing works in Spearman's correlation. It can be easily seen that the proposed ModelGiF yields significantly superior performance to existing methods, improving the SOTA Spearman's correlation from $0.777$ to $0.835$. Furthermore, all the three types of reference points produce Spearman's correlation more than $0.83$, which implies that the proposed method is quite robust to the choice of the reference points. To make a more comprehensive study, we test ModelGiF with varying number of reference points. The correlation curves are depicted in Figure~\ref{fig:taskonomy}e, where different types of reference points are compared. It can be seen that as the number of reference points increases, the correlation steadily grows. Surprisingly, with only $16$ points, ModelGiF alreadly achieves comparable performance to RSA~\cite{dwivedi2019representation} in Spearman's correlation.  It is attractive as the computation overhead of the proposed method grows linearly with the the number of reference points. The results indicate that we can safely reduce the computation cost by decreasing the number of reference points, and we can also strive for higher performance by adding more reference points, which makes ModelGiF flexible and applicable in a variety of scenarios. Another conclusion we can draw from Figure~\ref{fig:taskonomy}e is that the adversarial samples generally yields superior performance to other reference points, which gives implications of future work to strive for higher performance of task relatedness estimation. 

In Figure~\ref{fig:taskonomy}f, we make comparisons between some different implementations of ModelGiF. ``ModelGiF with random baselines'' denotes $\mathbf{x}_0$ in Eqn.~\ref{eq:prop-gif-curve} is fixed to be zero, and ``ModelGiF with random baselines'' denotes $\mathbf{x}_0$ is randomly sampled. We also include Integrated Gradient~(IG, with zero as the baseline) and IG with random baselines for comparisons. Albeit bearing some similarity with IG, it can be seen that ModelGiF with zero baselines significantly outperforms these baselines in most cases.  ModelGiF with random baselines, however, is more promising when the number of reference points becomes sufficiently larger.

\subsection{Application: Intellectual Property Protection}
We evaluate the performance of ModelGiF for IP protection against different model stealing attacks. To make fair comparisons with SOTA methods, we follow the experimental settings of~\cite{guan2022you} to conduct our experiments. 

\vspace{0.5em}
\noindent \textbf{Experimental Setup.}
Five categories of stealing attacks are considered here to test the protection performance. including finetuning, pruning, transfer learning, model extraction and adversarial model extraction. For finetuning, Finetune-L denotes fine-tuning only the last layer and leaves the other layers unchanged. Finetune-A fine-tunes all the layers in the model. For transfer learning, the CIFAR10 model is transferred to CIFAR10-C and CIFAR100. The tiny-ImageNet model (trained with the front 100 labels in Tiny-ImageNet) is transferred to the 100 labels left behind in tiny-ImageNet dataset. In model extraction, the victim model can be extracted in two manners: probability-based model extraction~(Extract-P), and label-based model extraction~(Extract-L). However, the attacker can evade the detection by applying adversarial training after the label-based model extraction~\cite{lukas2019deep}. In our experiment, adversarial model extraction~(Extract-adv) adopts the predicted label to evade the detection by adversarial training.

\vspace{0.5em}
\noindent \textbf{Models and Competitors.}
Different IP protection methods are evaluated on most of the common model architectures, including VGG~\cite{simonyan2014very}, ResNet~\cite{he2016deep}, DenseNet~\cite{huang2017densely} and MobielNet~\cite{sandler2018mobilenetv2}. To demonstrate the superiority of the proposed ModelGiF, we make comparisons with SOTA IP protection methods, including IPGuard~\cite{cao2021ipguard}, CAE~\cite{lukas2019deep}, EWE~\cite{jia2021entangled} and SAC~\cite{guan2022you}. IPGuard and CAE utilizes the transferability of adversarial examples and test the attack success rate of these adversarial examples on the suspect models. A model will be recognized as a stolen model if its attack success rate is larger than a threshold. SAC propose to leverage the pairwise relationship between samples as the model fingerprint. EWE, on the contrary, trains the source model on backdoor data and leaves the watermark in the model. Please refer to~\cite{guan2022you} for more detailed experimental settings.

\vspace{0.5em}
\noindent \textbf{Results.} To validate the effectiveness and the superiority of the proposed ModelGiF, we conduct experiments on different datasets for the defender and the attacker. We we leverage AUC-ROC curve and use AUC value between the fingerprinting scores of the irrelevant models and the stolen models to measure the fingerprinting effectiveness. Results are listed in Table~\ref{tab:ip-protection}. It can be seen that the proposed ModelGiF, although not tailored for IP protection, achieve superior performance in all the attack scenarios. The AUC value is $1.0$ across all attacks~(including the challenging ``Extract-adv'' attack), which means it perfectly recognize all the attacks to protect the IP, outperforming existing SOTA approaches like SAC~\cite{guan2022you}. We acknowledge the existing benchmark for IP protection methods is not sufficient large and challenging to provide thorough comparisons between IP methods. However, the superiority of ModelGiF to SOTA methods on this benchmark still provides us a strong confidence of the proposed method for quantifying model functional distance.    

\begin{figure*}[htbp]
  \centering
  \includegraphics[width=0.48\linewidth]{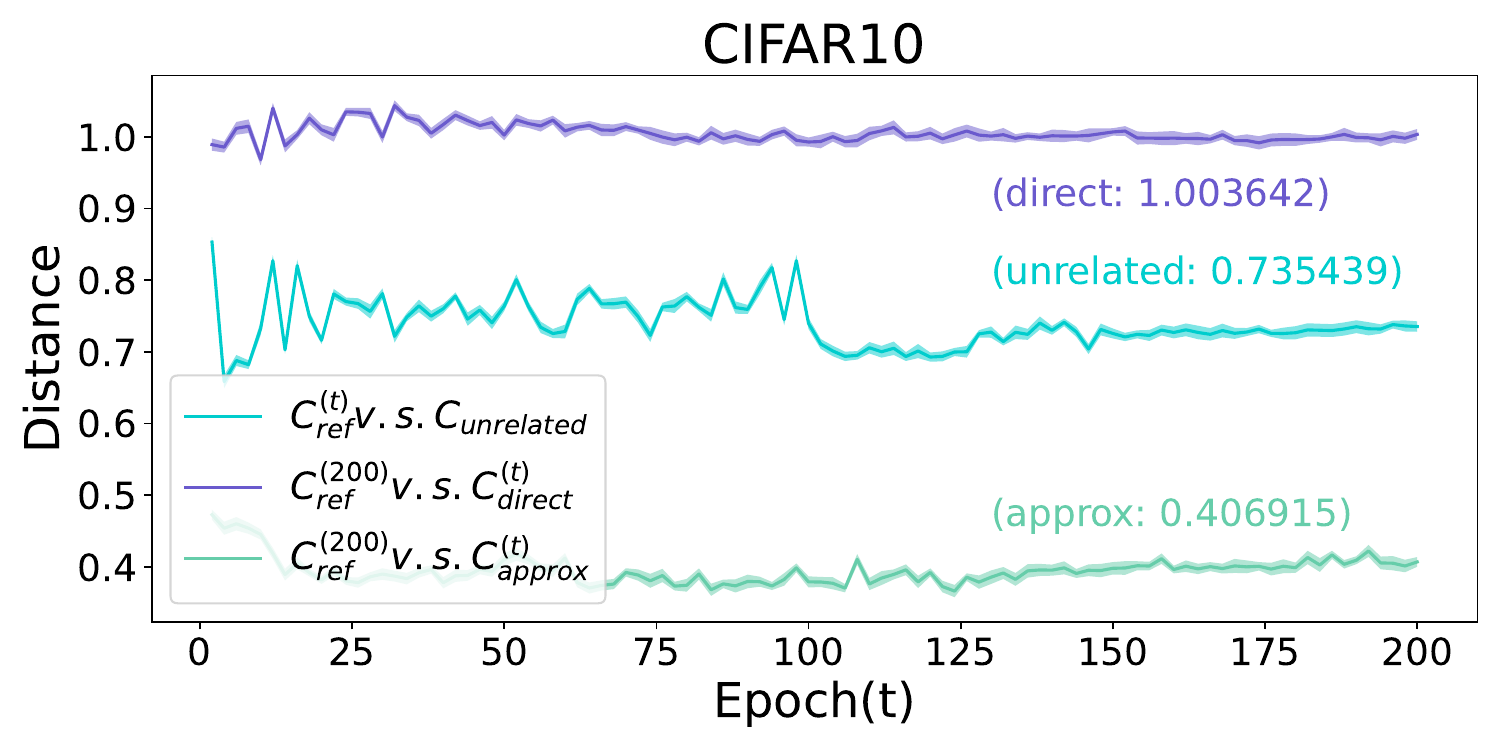}
  \includegraphics[width=0.48\linewidth]{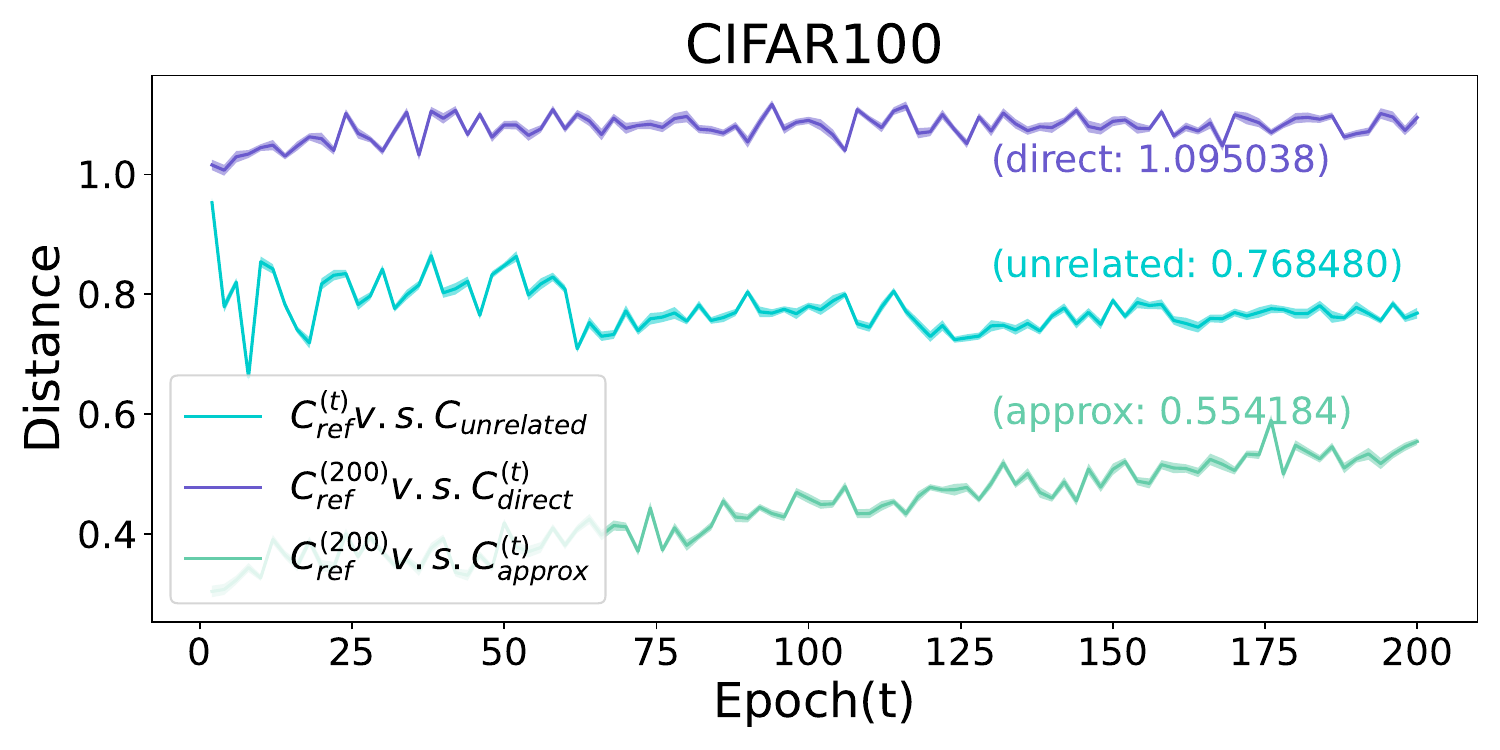}
  \caption{Cosine distances between the reference classifier $C_{ref}$ and unrelated classifier $C_{unrelated}$, the directly unlearned classifier $C_{direct}$ and the approximately unlearned classifier $C_{approx}$.}
  \label{fig:unlearning}
\end{figure*}
\subsection{Application: Model Unlearning Verification}
Machine unlearning studies how we can efficiently delete data points used to train models without retraining from scratch~\cite{bourtoule2021machine}. In this section, we demonstrate the effectiveness of the proposed ModelGiF for model unlearning verification, \ie, ModelGiF can distinguish the models trained with and without the data points that is requested to be unlearned.

\vspace{0.5em}
\noindent \textbf{Experimental Setup.}
Following the experimental settings of~\cite{jia2021zest}, we training four classifiers to evaluate ModelGiF for model unlearning verification. The first classifier is called the \textit{reference classifier} $C_{ref}^{(t)}$, where $t$ denotes that the classifier is obtained after $t$ epochs training. The reference classifier serves as the original classifier trained on all the training data, including the data points which are requested to be deleted later. The second classifier is called the \textit{unrelated classifier}, which is another classifier trained on all the training data, but from different initialization. The third classifier is called the \textit{exactly unlearned classifier}, which is trained on the remaining data after removing the data points requested to be deleted. Note that the exactly unlearned classifier is trained from scratch and has never seen those data points which are requested to be removed, and it thus can be seen as the exactly unlearned
classifier~\cite{bourtoule2021machine}. The last classifier is called \textit{approximately unlearned classifier}, which is obtained by directly optimizing the original reference classifier to
remove the knowledge learned from those data points requested to be unlearned~\cite{graves2021amnesiac}. We compare the ModelGiFs of the unrelated classifier, the exactly unlearned classifier and the approximately unlearned classifier to that of the reference classifier. The goal is to test whether the proposed ModelGiF can recognize the exactly unlearned classifier from the unrelated classifier. 

\vspace{0.5em}
\noindent \textbf{Experimental Details and Results.} Experiments are conducted on CIFAR10 and CIFAR100. On CIFAR10, all classifiers are implemented by ResNet20. On CIFAR100, all classifiers are implemented by ResNet50. We randomly sample $128$ data points from the training data to be unlearned, and use these data as the reference points to compute the distance between ModelGIFs of these classifiers.  Experimental results are provided in Figure~\ref{fig:unlearning}. It can be seen that with the proposed ModelGiF, exactly unlearned classifier gets significantly higher distance with the reference classifier than the unrelated classifier. It implies that ModelGiF can also be used as a tool to verify the unlearning performance  of existing unlearning methods. Another observation from Figure~\ref{fig:unlearning} is that the distance between the approximately unlearned classifier and the reference classifier is much lower than unrelated classifier, which implies that existing approximately unlearning methods can not delete the data from the model thoroughly. However, as the training for unlearning continues, the information can be gradually forgotten, as implied by the increasing distance shown in Figure~\ref{fig:unlearning}. These results also accord with prior finding that continual learning easily lead to catastrophic forgetting problem~\cite{Parisi2018ContinualLL}, which validate the rationality of the results from the proposed ModelGiF.

 \section{Conclusion and Future Work}
In this work, we propose ModelGiF to quantify model functional distance. The main assumption underlying ModelGiF is that each pre-trained deep model uniquely determines a ModelGiF over the input space. The distance between models can thus be measured by the similarity between their ModelGiFs. We apply the proposed ModelGiF to task relatedness estimation, intellectual property protection, and model unlearning verification. Experimental results demonstrate the versatility of the proposed ModelGiF on these tasks, with significantly superiority performance to state-of-the-art competitors.
There are several directions for future work with the proposed ModelGiF. For example, exploring more scenarios where ModelGiF can be applied. Another interesting research direction is proposing more informative reference points to extract stronger representations of ModelGiF. Finally, how to further speed up the computation of the proposed method is also important to make it easier to use.

\noindent\textbf{Acknowledgement.} This work is funded by the National Key Research and Development Project (Grant No: 2022YFB2703100), National Natural Science Foundation of China (62106220, 61976186, U20B2066), and Ningbo Natural Science Foundation (2021J189).

{
\bibliographystyle{ieee_fullname}
\bibliography{egbib}
}

\end{document}